\documentclass[conference]{IEEEtran}
\IEEEoverridecommandlockouts
\usepackage{cite}
\usepackage{amsmath,amssymb,amsfonts}
\usepackage{algorithmic}
\usepackage{breqn}
\usepackage{graphicx}
\usepackage{hyperref}
\hypersetup{
    colorlinks=true,
    linkcolor=blue,
    filecolor=magenta,      
    urlcolor=blue,
    pdftitle={Overleaf Example},
    pdfpagemode=FullScreen,
    }
\usepackage{textcomp}
\usepackage{xcolor}
\newcommand\mdoubleplus{\mathbin{+\mkern-10mu+}}
\def\BibTeX{{\rm B\kern-.05em{\sc i\kern-.025em b}\kern-.08em
    T\kern-.1667em\lower.7ex\hbox{E}\kern-.125emX}}
\begin{document}

\title{Ensemble of Task-Specific Language Models for Brain Encoding}

\author{
\IEEEauthorblockN{Arvindh Arun}
\IEEEauthorblockA{Precog\\
\textit{IIIT Hyderabad}}
\and
\IEEEauthorblockN{Jerrin John Thomas}
\IEEEauthorblockA{Language Technologies Research Center\\
\textit{IIIT Hyderabad}}
\and
\IEEEauthorblockN{Sanjai Kumaran P}
\IEEEauthorblockA{Centre for Visual Information Technology\\
\textit{IIIT Hyderabad}}}

\maketitle

\begin{abstract}
Language models have been shown to be rich enough to encode fMRI activations of certain Regions of Interest in our Brains. Previous works have explored transfer learning from representations learned for popular natural language processing tasks for predicting brain responses. In our work, we improve the performance of such encoders by creating an ensemble model out of 10 popular Language Models (2 syntactic and 8 semantic). We beat the current baselines by 10\% on average across all ROIs through our ensembling methods.
\end{abstract}

\section{Introduction}
\noindent
Brain encoding is the process of converting textual, visual, or any sensory information to neural activity patterns. Language models can be used to predict neural activity. Research has been conducted on the potential of task-specific fine-tuned language models to predict fMRI brain activity. Language models (LMs) trained on large corpora often tend to express cognitive understanding. Transformers are a class of LMs that do it the best and are hard to probe and interpret. However, further scope exists for incorporating information from multiple task-specific models to improve encoding accuracy. Our work investigates the effectiveness of combining multiple task-specific models to predict fMRI brain activity across different brain regions.

This approach uses encoding models based on task features to predict brain activity across various Regions of Interest in the brain (ROIs). However, rather than relying solely on task features, we seek to enhance encoding accuracy by incorporating information from multiple task-specific models. By combining the insights from these models, we may improve our predictions' accuracy and gain a more comprehensive understanding of the underlying neural processes.

We develop an effective ensemble of task-specific language models that can improve the accuracy and efficiency of brain encoding. We evaluate the models and discuss the implications of our findings for the field of cognitive neuroscience. Our work contributes to the ongoing efforts to understand the neural processes underlying cognition and their relationship with language. By leveraging language models' capabilities, we can better understand how the brain processes and represents information during various cognitive tasks. Our findings have implications for developing more accurate and robust encoding models for fMRI data analysis and provide insights into the relationship between language and brain activity.

\section{Related Work}

Gauthier and Levy (2019)\cite{b2} investigate the robustness of human brain representations of sentence understanding. They compare various sentence encoding models on a brain decoding task where the sentence the participant sees must be predicted from the fMRI signal evoked by the sentence. They use a pre-trained BERT architecture as a baseline and fine-tune it on various natural language understanding tasks to determine which leads to improvements in brain-decoding performance. However, they find that none of the tested sentence encoding tasks yield significant increases in brain decoding performance. Further task ablations and representational analyses reveal that tasks producing syntax-light representations significantly improve brain decoding performance. These results offer constraints on the space of NLU models that best account for human neural representations of language and suggest limitations on decoding fine-grained syntactic information from fMRI human neuroimaging.

Schrimpf et al. (2021)\cite{b3} report on integrating recent artificial neural networks from machine learning with human recordings during language processing. The study finds that the most powerful models predict neural and behavioral responses across different datasets up to noise levels. Furthermore, their study shows that models that better predict the next word in a sequence also better predict brain measurements. These findings suggest that predictive processing plays a fundamental role in shaping language comprehension mechanisms in the brain. The study provides evidence for a neurally-mechanistic account of how meaning might be extracted from language, which has long been lacking in the field of cognitive neuroscience.

Oota et al. (2022)\cite{b1} explore the efficacy of task-specific learned Transformer representations for predicting brain responses in two diverse datasets: Pereira and Narratives. They use encoding models based on task features learned from ten popular natural language processing tasks, including both syntactic and semantic tasks, and find that features from coreference resolution, NER, and shallow syntax parsing explain greater variance for the reading activity, while tasks such as paraphrase generation, summarization, and natural language inference show better encoding performance for the listening activity. Their experiments provide insights into the cognitive aspects of language processing in the human brain. Our work is an extension of their work.

\setlength{\tabcolsep}{2.5pt}
\begin{table}[!ht]
\scriptsize
\centering
\begin{tabular}{|l|c|c|c|c|c|c|c| c |c |}
\hline
 ROIs & \multicolumn{2}{c|}{Language} & \multicolumn{5}{c|}{Vision} & \multicolumn{1}{c|}{DMN} & \multicolumn{1}{c|}{Task Positive} \\ \cline{1-10} 
Subj& LH& RH & Body & Face& Object & Scene & Vision& RH & LH  
\\ \hline 
P01&5265&6172&3774&4963&8085&4141&12829&17190&35120 \\
M02&4930&5861&3873&4782&7552&3173&11729&15070&30594 \\
M04&5906&5401&3867&4803&7812&3602&12278&18011&34024 \\
M07&5629&5001&4190&4993&8617&3721&12454&17020&30408 \\
M15&5315&6141&4112&4941&8323&3496&12383&15995&31610 \\
\hline
\end{tabular}\\
\caption{Number of Voxels in each ROI in the Pereira Dataset. LH - Left Hemisphere. RH - Right Hemisphere.}
\label{tab:pereira_stats}
\end{table}

\section{Dataset}
We use the Pereira dataset, which contains brain responses from subjects reading sentences. We utilize the data from sentence-based experiments (experiments 2 and 3) conducted by Pereira et al. (2018) \cite{b4}. A total of 627 sentences from 48 topics, presented to five subjects, were analyzed. These sentences were part of 168 passages, each containing 3-4 sentences. Our analysis focuses on nine brain ROIs belonging to four brain networks: the Default Mode Network (DMN), Language Network, Task Positive Network (TP), and Visual Network. The DMN is linked to semantic processing, while the Language Network is associated with language processing, understanding, word meaning, and sentence comprehension. The TP network is related to attention and salience information, and the Visual Network is responsible for visual object processing and object recognition.

\section{Methodology}
\subsection{Baselines}
For the baselines, one encoder model is used for each task. Feature spaces, describing stimulus sentences, are extracted and used to predict brain activity in the encoding model. The encoder is trained using Ridge regression for several NLP tasks. A model is trained for each subject and each ROI. For the data. train-test split was done in the ratio 4:1.

\begin{table}[h]
\begin{tabular}{|l|l|}
\hline
\multicolumn{1}{|c|}{\textbf{Task}} & \multicolumn{1}{c|}{\textbf{Model Name}} \\ \hline
General                             & \href{https://huggingface.co/bert-base-uncased}{bert-base-uncased}                        \\ \hline
Coreference Resolution              & \href{https://huggingface.co/nielsr/coref-bert-base}{coref-bert-base}                          \\ \hline
Named Entity Recognition            & \href{https://huggingface.co/dslim/bert-base-NER}{bert-base-NER}                            \\ \hline
Natural Language Inference          & \href{https://huggingface.co/sentence-transformers/bert-base-nli-mean-tokens}{bert-base-nli-mean-tokens}                \\ \hline
Paraphrase                          & \href{https://huggingface.co/bert-base-cased-finetuned-mrpc}{bert-base-cased-finetuned-mrpc}           \\ \hline
Question Answering                  & \href{https://huggingface.co/docs/transformers/model_doc/bert#transformers.BertForQuestionAnswering}{bert-base-cased-squad2}                   \\ \hline
Sentiment Analysis                  & \href{https://huggingface.co/barissayil/bert-sentiment-analysis-sst}{bert-sentiment-analysis-sst}              \\ \hline
Semantic Role Labeling              & \href{https://huggingface.co/liaad/srl-en_mbert-base}{srl-en\_mbert-base}                       \\ \hline
Shallow Syntax                      & \href{https://huggingface.co/vblagoje/bert-english-uncased-finetuned-chunk}{bert-english-uncased-finetuned-chunk}     \\ \hline
Summarization                       & \href{https://huggingface.co/lidiya/bart-base-samsum}{bart-base-samsum}                         \\ \hline
Word Sense Disambiguation           & \href{https://github.com/BPYap/BERT-WSD}{BERT-base-augmented}                      \\ \hline
\end{tabular}\\
\caption{Task-specific models used}
\end{table}

We encode fMRI data using several natural language processing (NLP) tasks. These tasks include coreference resolution (CR), paraphrase detection (PD), summarization (Sum), named entity recognition (NER), natural language inference (NLI), question answering (QA), sentiment analysis (SA), semantic role labeling (SRL), syntactic structure approximation (SS), and word sense disambiguation (WSD). Each task serves a specific purpose in analyzing and understanding language, such as identifying entities and their relationships, detecting sentiment, and determining the meaning of words in context.

\begin{itemize}
    \item Coreference Resolution (CR): finds all expressions in a text that refer to the same entity.
    \item Paraphrase Detection (PD): rewords a given passage in shorter or different words while preserving its meaning.
    \item Summarization (Sum): selects a few important sentences from a document or paragraph.
    \item Named Entity Recognition (NER): detects named entities, such as person names, location names, and company names, in a given text.
    \item Natural Language Inference (NLI): investigates the entailment relationship between two texts, a premise, and a hypothesis.
    \item Question Answering (QA): selects an answer from a set of candidate answers given a passage and a question.
    \item Sentiment Analysis (SA): determines whether a piece of text is positive, negative, or neutral.
    \item Semantic Role Labeling (SRL): assigns labels to words or phrases in a sentence that indicate their semantic role in the sentence, such as that of an agent, goal, or result.
    \item Shallow Syntax Parsing (SS): approximates the phrase-syntactic structure of sentences.
    \item Word Sense Disambiguation (WSD): determines which sense or meaning of a word is activated by its use in a particular context.
\end{itemize}

\begin{figure*}[ht]
\centerline{\includegraphics[width=2\columnwidth]{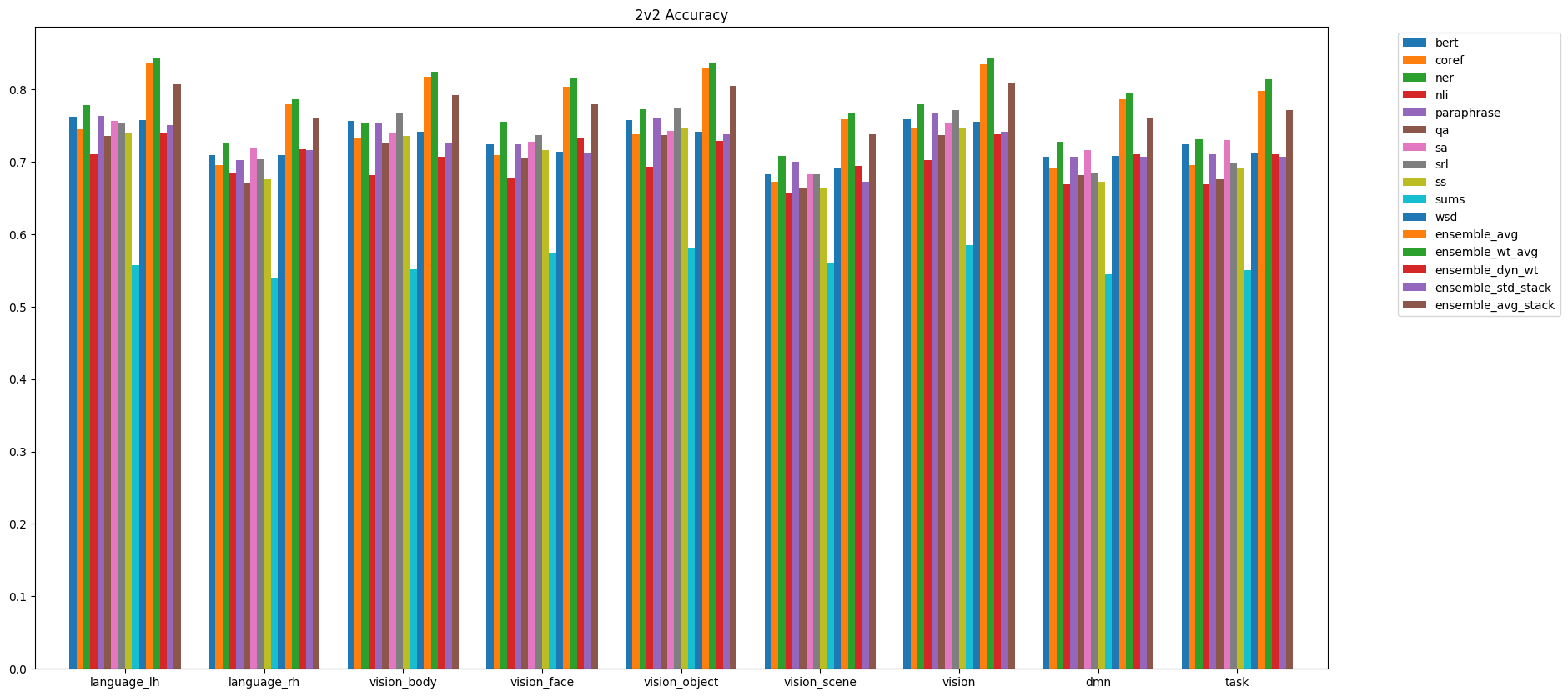}}
\label{fig}
\centerline{\includegraphics[width=2\columnwidth]{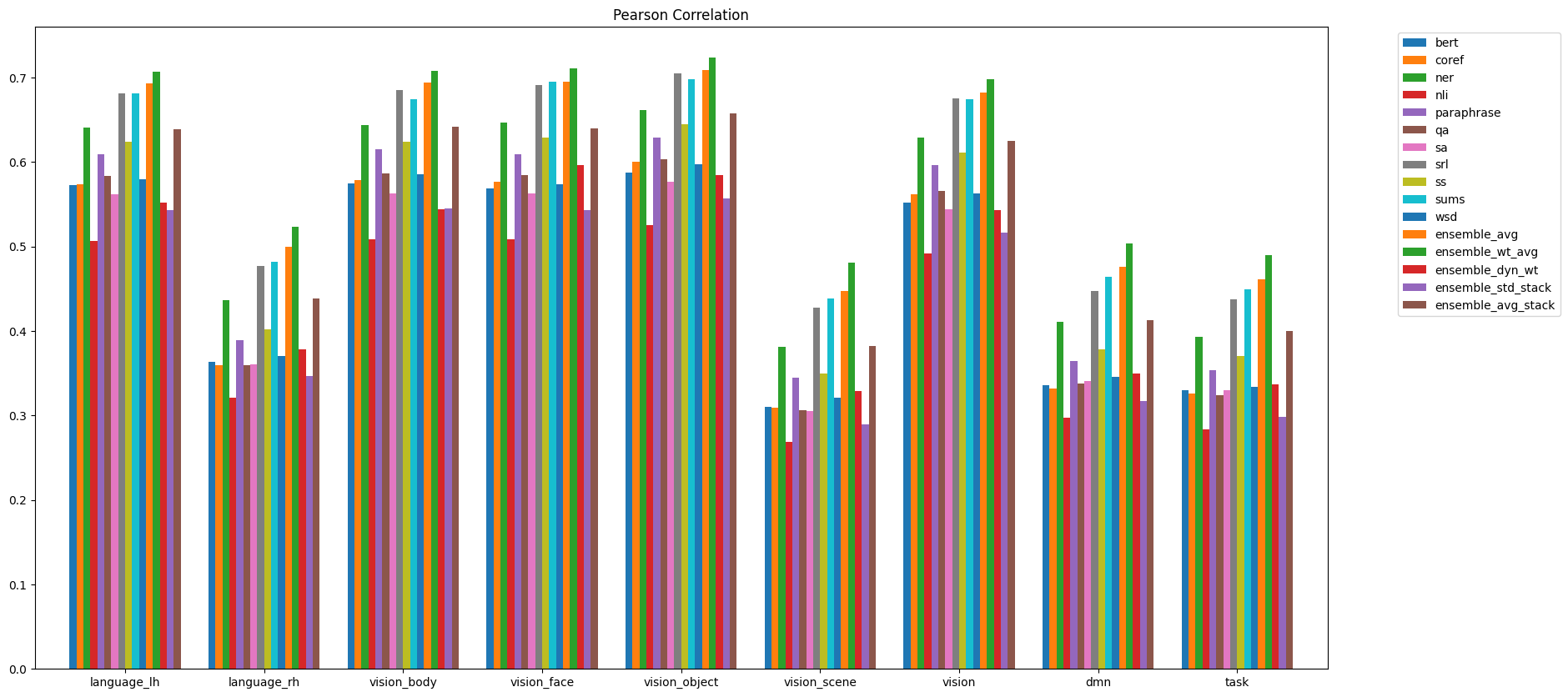}}
\caption{2v2 Accuracty and Pearson correlation}
\end{figure*}

\subsection{Ensembling Methods}
We hypothesize that combining multiple task-specific models for a single ROI will be able to generate better representations as each ROI is not specifically dedicated to only a sole task. This serves as our primary motivation to try out different ensembling methods with the task-specific models' outputs. For reproducibility, we have made the code public \footnote{Code - \url{https://github.com/jr-john/ensemble_brain_encoders}}.

We explore multiple ensemble methods ranging from simple centroid calculation (Averaging) to learning a meta-model to combine the representations. Let $n = 11$ correspond to the number of task-specific Language Models we consider for our experiments. Each of the $n$ embeddings generated is of the same dimension - $(768,)$. Let $u_{i}$ correspond to the embedding vector generated by the $i$-th LM and $u_{f}$ to the final ensemble embedding.

\subsubsection{Average}
We take the dimension-wise average of embeddings generated by all the task-specific pre-trained LMs. Geometrically, this corresponds to the centroid of the polyhedron formed by all the $n$ points in the embedding representation space.
\begin{dmath*}
    u_f = \dfrac{\sum_{i=1}^{n} u_{i}}{n}
\end{dmath*}
The sum here is dimension-preserving, i.e., it is taken across each of the $768$ dimensions.

\subsubsection{Weighted Average}
The weights were decided heuristically, by looking at which task has higher predictivity for a particular ROI and the similarity between the tasks. We weigh the outputs of different LMs with these weights. $w_{i}$, a scalar, corresponds to the weight assigned to the $i$-th LM's embedding vector. Operation $(\cdot)$ corresponds to the vector dot product.
\begin{dmath*}
    u_f = \dfrac{\sum_{i=1}^{n} w_{i} \cdot u_{i}}{n}
\end{dmath*}
where the weight $w_{i}$ is defined using the power mean of accuracy $x_{i}$ for each task-specific model as,
\begin{dmath*}
    w_i = \left(\dfrac{1}{n} \sum_{i=1}^{n} x_{i}^p\right)^{\dfrac{1}{p}}
\end{dmath*}
where $p$ is a factor that controls the influence of weights. We experiment with various $p$ values. When the p-value is close to 0, the weights assigned to the embeddings with the larger weights dominate while when the p-value is large, the weights assigned to the embeddings with the smaller weights dominate. We observed that increasing the p-value, improves the model performance and it keeps on improving till p = 5, and then starts decreasing.

\subsubsection{Dynamic Weights}
Instead of statically defining the weights using heuristics, the weights are learned from the training data. They are not predetermined, rather training adjusts the weights while optimizing the loss function. The weights determine the importance of each model output embedding, that is, it tells if the task is important for a particular ROI. The loss function used is MSE loss and optimizer is Adam optimizer. The weights are constrained to be positive and are normalized to unity.

\subsubsection{Stacking with PCA}
Initially, the aim was to stack all the weak learner output embeddings and pass it to the meta-model. However, this required huge amounts of memory and was not feasible which made dimensionality reduction a necessity. We pass all the outputs through PCA to generate representative low-dimensional embeddings. We then concatenate them together, passing them through a meta-learner to generate the final predictions. Let $f$ be a function that denotes the meta-model,
\begin{dmath*}
    u_f = f\left(\mdoubleplus_{i=1}^{n} \text{PCA}(u_{i})\right)
\end{dmath*}

\subsubsection{Stacking with Average}
PCA leads to some amount of information loss. Discarding the dimensions with low variance will leads to loss of some information. Applying PCA on the different base model outputs, leads to each one having very different distributions. Therefore, instead of PCA and then concatenation, we take an average of all the outputs and then pass it through a meta-learner.
\begin{center}
    $u_f = f\left(\dfrac{\sum_{i=1}^{n} u_{i}}{n}\right)$
\end{center}

\subsection{Evaluation}
We evaluate our models using popular brain encoding evaluation metrics - 2v2 Accuracy and Pearson Correlation. Let $N$ be the number of samples given a subject and a brain region. Let $\{Y_i\}_{i=1}^N$ and $\{\hat{Y}_i\}_{i=1}^N$ denote the actual and predicted voxel value vectors for the $i$-th sample. Thus, $Y \in \mathbb{R}^{N \times V}$ and $\hat{Y} \in \mathbb{R}^{N \times V}$ where $V$ is the number of voxels in that region.

\noindent
\textbf{Pearson Correlation (PC)} is defined as, 
\begin{dmath*}
    \text{PC} = \frac{1}{N} \sum_{i=1}^{N} \operatorname{corr}(Y_i, \hat{Y_i})
\end{dmath*}

\noindent
\textbf{2v2 Accuracy} is calculated as,
\begin{dmath*} 
2V2\text{Acc} = \frac{1}{\binom{N}{2}}\sum_{i=1}^{N-1}\sum_{j=i+1}^{N} I\left[\cos(D(Y_i, \hat{Y_i}) + \cos(D(Y_j, \hat{Y_j})) < \cos(D(Y_i, \hat{Y_j})) + \cos(D(Y_j, \hat{Y_i}))\right]
\end{dmath*}
For both metrics, a higher value indicates better encodings.

\section{Results}
The results are shown in Figure 1. The approach of weighted average using the heuristic weights performs the best out of all, followed by simple averaging and then stacking after averaging. These three approaches clearly outperform the baselines by about 10\%. This shows that averaging across output embeddings from task-specific models is a very good way to ensemble the models. The approach of stacking with PCA did not work very well because the dimensionality reduction lost some information and also changed the distribution. As for the ensemble method of finding the weights dynamically, the most probable issue is that weights were overfitting on the training data since the number of stimulus sentences is very less per subject. The results are available in the analysis notebook in the codebase (link provided above).

\section{Conclusion}
Thus we were able to create several ensembles of task-specific language models. Most of them outperformed the baselines in terms of encoder accuracy and performance. The method of averaging seemed to be the best ensemble technique. Also, the weights from approaches 2 and 3 give insight into which task is involved in each ROI. If a particular feature gives good predictivity for a particular ROI, then information for that specific is most likely to be encoded in that ROI. Thus, the weights tell which tasks are important for which ROI and which ROI has better predictivity for each task.

\vspace{12pt}


\begin{thebibliography}{00}
\bibitem{b1} Subba Reddy Oota, Jashn Arora, Veeral Agarwal, Mounika Marreddy, Manish Gupta, and Bapi Surampudi. 2022. Neural Language Taskonomy: Which NLP Tasks are the most Predictive of fMRI Brain Activity?. In Proceedings of the 2022 Conference of the North American Chapter of the Association for Computational Linguistics: Human Language Technologies, pages 3220–3237, Seattle, United States. Association for Computational Linguistics.

\bibitem{b2} Jon Gauthier and Roger Levy. 2019. Linking artificial and human neural representations of language. In Proceedings of the 2019 Conference on Empirical Methods in Natural Language Processing and the 9th International Joint Conference on Natural Language Processing (EMNLP-IJCNLP), pages 529–539, Hong Kong, China. Association for Computational Linguistics.

\bibitem{b3} Schrimpf, M., Blank, I. A., Tuckute, G., Kauf, C., Hosseini, E. A., Kanwisher, N., Tenenbaum, J. B., \& Fedorenko, E. (2021). The neural architecture of language: Integrative modeling converges on predictive processing. Proceedings of the National Academy of Sciences, 118(45), e2105646118. https://doi.org/10.1073/pnas.

\bibitem{b4} Francisco Pereira, Bin Lou, Brianna Pritchett, Samuel Ritter, Samuel J Gershman, Nancy Kanwisher, Matthew Botvinick, and Evelina Fedorenko. 2018. Toward a universal decoder of linguistic meaning from brain activation. Nature communications, 9(1):1–13.
\end{thebibliography}
\end{document}